\pdfoutput=1

\documentclass[11pt]{article}

\usepackage[]{acl}

\usepackage{times}
\usepackage{latexsym}
\usepackage{amsmath}

\usepackage{graphicx}

\usepackage[T1]{fontenc}

\usepackage[utf8]{inputenc}

\usepackage{microtype}

\usepackage{algorithm}
\usepackage{algorithmic}
\usepackage{mathrsfs}
\usepackage{amsmath}
\usepackage{graphicx}
\usepackage{booktabs}
\usepackage{array}
\usepackage{boldline}
\usepackage{multirow,tabularx}

\title{AllWOZ: Towards Multilingual Task-Oriented Dialog Systems for All}

\author{Lei Zuo,  Kun Qian,  Bowen Yang, Zhou Yu \\
  Columbia University \\
  \texttt{\{lz2771, kq2157, by2299, zy2461\}@columbia.edu}}

\begin{document}
\maketitle
\begin{abstract}

A commonly observed problem of the state-of-the-art natural language technologies, such as Amazon Alexa and Apple Siri, is that their services do not extend to most developing countries' citizens due to language barriers. Such populations suffer due to the lack of available resources in their languages to build NLP products. This paper presents AllWOZ, a multilingual multi-domain task-oriented customer service dialog dataset covering eight languages: English, Mandarin, Korean, Vietnamese, Hindi, French, Portuguese, and Thai. Furthermore, we create a benchmark for our multilingual dataset by applying mT5~\cite{xue-etal-2021-mt5} with meta-learning \cite{Finn2017ModelAgnosticMF}. 

\end{abstract}

\section{Introduction}
\label{sec:intro}
Task-oriented dialog systems are crucial for business solutions. While task-oriented dialog systems have made tremendous success in English, there is still a pressing urgency to build systems that can serve 6,900 other languages all over the world to enable universal technology access \cite{Ruder2019ASO, Aharoni2019MassivelyMN, Arivazhagan2019MassivelyMN}. 94\% of the world's population do not have English as their first language, and 75\% do not speak English at all. Most developing countries' citizens cannot benefit from state-of-the-art language technologies due to language barriers.

Building dialog systems for most languages is challenging due to a lack of data. Automatic translation is a powerful tool to generate more resources. However, the state-of-the-art translations still suffer from low fluency and coherence. Moreover, they have difficulties dealing with the mentioned entities in the dialog, which is essential in serving the functional purposes of task-oriented dialog systems.

\begin{figure}
\centering
\includegraphics[width=\columnwidth]{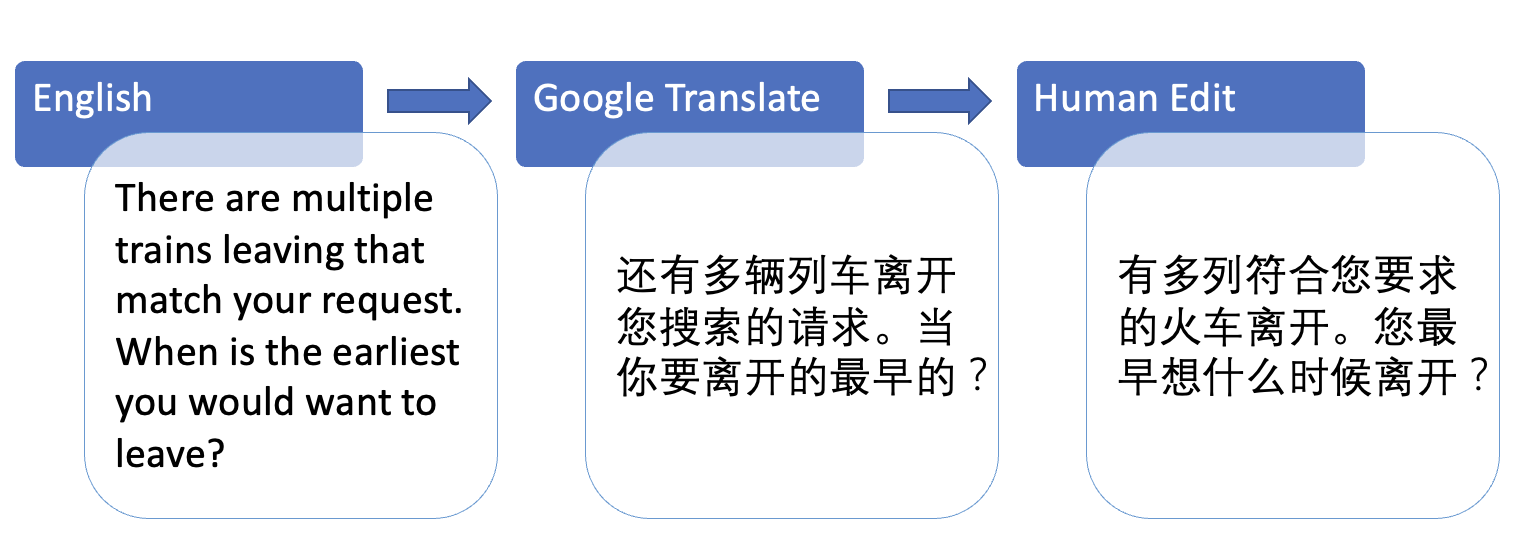}
\caption{Three stages of our data collection: Data selection, machine translation and human correction.}
\label{fig_translate}
\end{figure}

To facilitate the development of multilingual task-oriented dialog systems, we create a new dataset AllWOZ based on MultiWOZ \cite{Budzianowski2018MultiWOZA, zang2020multiwoz}. AllWOZ is a multilingual multi-domain task-oriented dialog dataset with intent and state annotation. It has eight languages across various language families: English, Mandarin, Korean, Vietnamese, Hindi, French, Portuguese and Thai. We will extend the dataset to more than 20 languages in our future work.

Many languages have similarities in syntax and vocabulary, and multilingual learning approaches that leverage the shared structure of the input space have proven to be effective in alleviating data sparsity. In this work, we applied a meta-learning training schema for multilingual adaptation to take advantage of shared language structures.

Our contributions are as follows: (1) We collect a new dataset, AllWOZ, for multilingual task-oriented dialog systems. (2) With the experiments, we conclude that few-shot learning could improve the model performance on our dataset. We would make our dataset and models public.

\section{Related Work}
Early work in this direction focused on individual tasks, such as grammar induction \cite{Ruder2019ASO, snyder-etal-2009-unsupervised}, part-of-speech (POS) tagging \cite{tackstrom-etal-2013-token}, parsing \cite{mcdonald-etal-2011-multi}, and text classification \cite{klementiev-etal-2012-inducing}. General-purpose multilingual representation learning has gained increasing attention during the past few years. Approaches that are applicable to multiple tasks have been researched on both word-level \cite{Mikolov2013EfficientEO, faruqui-dyer-2014-improving, artetxe-etal-2017-learning} and sentence-level \cite{Devlin2019BERTPO, Lample2019CrosslingualLM}. However, previous work processed text within a short context window due to the lack of datasets with long texts. There is little publicly available dialog resource that contains a diverse set of languages. A multilingual multi-domain natural language understanding (NLU) dataset with Thai, English, and Spanish \cite{Schuster2019CrosslingualTL}. \cite{Mrksic2017SemanticSO} annotated only two additional languages in WOZ 2.0 and \cite{liu-etal-2019-zero} proposed a mixed-language training method for cross-lingual NLU and dialog state tracking (DST) tasks.

In terms of algorithms, \citet{Schuster2019CrosslingualTL} found that in low resource settings, multilingual contextual word representations produce better results than using cross-lingual static embeddings. This suggests that simply using pre-trained multilingual embedding, such as MASS \cite{Song2019MASSMS} and mBART \cite{Liu2020MultilingualDP}, which trained on auto-encoding objectives are not ideal for solving the dialog task. This prompts us to propose new algorithms that not only utilizes pretrained multilingual embedding, but also considers dialogue context information.

\section{Multilingual Dialog Collection}

To build a multilingual task-oriented dialog system, we collect a new dataset, AllWOZ, consisting of paired dialogs between different languages based on the MultiWOZ dataset.
We first carefully sample dialogs from MultiWOZ and then translate those dialogs into different languages with Google Translation Tool. To ensure the quality of the translation, we recruit native speakers for each language to correct the translation results.

\subsection{Data Selection} MultiWOZ~\cite{Budzianowski2018MultiWOZA} is the most popular task-oriented dialog dataset, covering seven domains and containing 10K+ dialogs. 
Many works devote effort to correcting and improving the dataset~\cite{eric-etal-2020-multiwoz, qian-etal-2021-annotation, Han2021MultiWOZ2A, Ye2021MultiWOZ2A}. 
We conduct translation jobs on the MultiWOZ 2.2~\cite{zang2020multiwoz} since it is the most widely-accepted version.
As mentioned in Sec.~\ref{sec:intro}, most languages lack dialog training data, so our goal is to build dialog models under few-shot settings. 
Therefore, we sample 100 dialogs (1476 turns) from the test set in total.
In order to maintain the prior knowledge of each domain, we keep the same domain distribution of the whole test set during sampling.
For example, as shown in Table~\ref{table_stats}, there are 38 out of 100 sampled dialogs involved in the attraction domain, and 399 out of 1000 dialogs involved in the attraction domain in the test set.
Those two ratios are very close.
Similarly, for all five domains, the ratio of the dialog number counted for sampled dialogs (left side of slash in Table~\ref{table_stats}) over the number for the whole test set (right side of slash) keeps consistent.
The same case happens when it comes to the turn number.
As for each domain, we expect the sampled dialogs to cover as much information as possible. 
So, during the sampling, we record the dialog state annotations of chosen dialogs and skip the dialog with similar annotations.
As shown in the last row of Table~\ref{table_stats}, the sampled dialog covers all possible slot types.

\begin{figure}[t]
\centering
\includegraphics[width=\columnwidth]{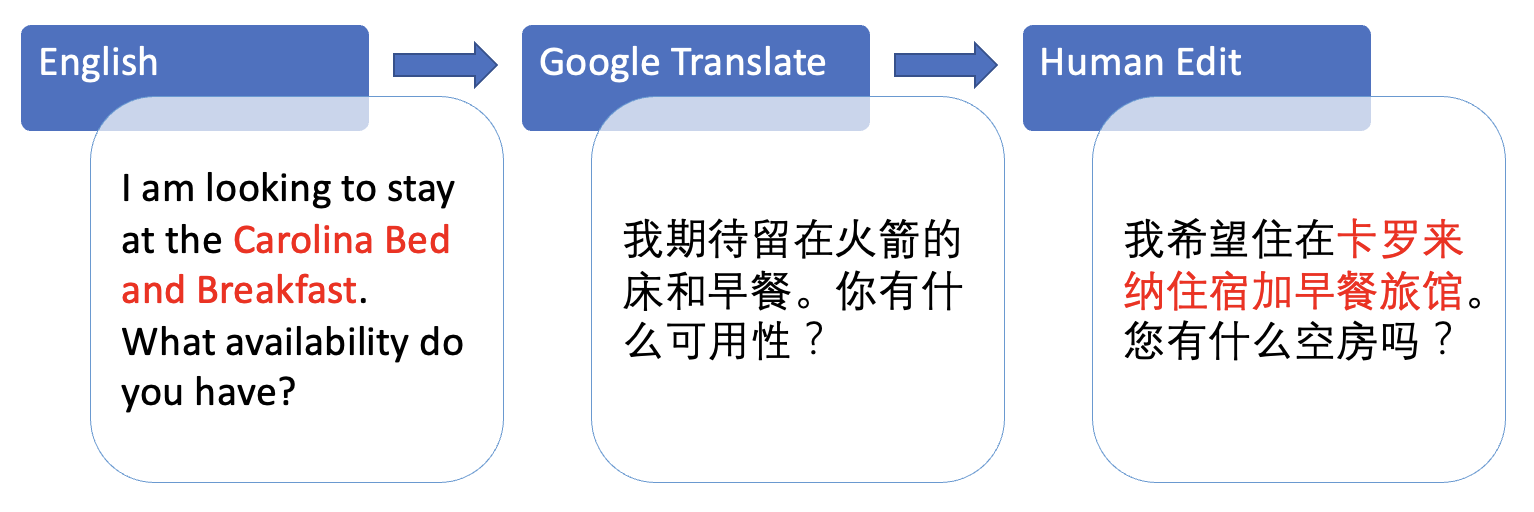}
\caption{
Data generated from machine translation are noisy when there are entities in the sentence. 
}
\label{fig_entity}
\end{figure}

\begin{table*}[t]
\centering
\small
\begin{tabular}{l|lllllll}
\toprule \hline
Num. of  & Attraction  & Hotel  & Restaurant & Taxi & Train \\ \hline
Dialog   &   38/399       &   40/395     &     39/445       &  13/198    &   55/489    \\
Turn     &  213/2433        &   254/2588     &   224/2867         &   38/640   &  331/2946     \\
Slot Type  &  3/3        &   10/10     &    7/7        &   4/4   &   6/6    \\
\hline \bottomrule
\end{tabular}
\caption{The statistics for the sampled 100 dialogs/whole test set in terms of dialog number, turn number and the number of different slot types over all five domains in test set. The sampled dialogs are evenly distributed across the five domains.}
\label{table_stats}
\end{table*}

\subsection{Machine Translation} In order to reduce the human workload of translation, we first utilize the Google Translation Tool to automatically translate both dialog utterances and dialog state annotations.
Fig.~\ref{fig_translate} shows an example of the translation flow. Machine-translated utterances are usually of low quality, mainly because some entity tokens like ``Carolina Bed and Breakfast'' is hard for machine to translate.

\subsection{Human Correction} To build a high-quality dataset, we recruit native speakers to correct the errors in the machine-translated utterances.
Our dataset currently covers eight languages: English, Mandarin, Korean, Vietnamese, Hindi, French, Portuguese and Thai.
For each language except English, we recruit a bilingual speaker to edit the machine-translated utterances based on the original English dialogs.
In addition to dialog utterances, we also require the translators to edit the machine translations of the dialog states (e.g. ``Carolina Bed and Breakfast'' in Fig.~\ref{fig_translate}), because tasks like dialog state tracking and end-to-end dialog generation require those states.
However, some entity tokens in dialog states have polysemy and the translation of the dialog states does not match the semantic meaning in the dialog utterance. 
For example, the token ``moderate'' refers to price by default in the dialog utterance. However, as an isolated token in dialog states, it is translated as ``mild''.
To ensure that all the translations of dialog states are natural and coherent, we ask the translator to translate all dialog states first.
Then, they should translate dialog utterances based on the dialog states' translations, in order to avoid inconsistency between utterances and slots.
If any state looks not coherent or natural in an utterance, translators are required to edit the translation of dialog states and translate all related utterances again.

\section{Experiments}

In this section, we introduce how we divide data for train, validation and evaluation, as well as the experiment setting.
\subsection{Data Partition}
For each language, the translated 100 dialogs are divided into three partitions, each with 40/10/50 dialogs. We first randomly sample a target language, then all the other seven languages are considered source languages. The experiments aim to explore whether the parallelism among source languages can help learn the target language under few-shot settings. For the source languages, we use the 40 dialogs as training data and 10 dialogs as validation data. And for the target language, since we focus on the few-shot learning, we utilize the partition of 10 dialogs as training data and 40 dialogs for validation. The remaining partition of 50 dialogs is used as the test set.
In order to achieve trustful results, we run each experiment for two times. Each time we randomly re-sample the data partition and in the table we report the average score, along with the standard deviation.




\subsection{Benchmark Models}
Inspired by the success of pre-trained multilingual model~\cite{Song2019MASSMS, Liu2020MultilingualDP, lin-etal-2020-pre}, we choose mT5~\cite{Xue2020mT5AM} as our backbone model, a multilingual pre-trained encoder-decoder language model. It is pre-trained on mC4~\cite{2019t5}, which covers 101 languages in total, including the 8 languages that we propose to translate. The experiment are conducted under the following settings: 

\begin{itemize}
  \item \textbf{Vanilla Training} The vanilla method is directly fine-tuning mT5 model with mixed dialogs from each source languages and then test on the target language.
  \item \textbf{Vanilla + English Pretrain} Inspired by the success of pre-trained language models, we first pre-train the model on the full-size MultiWOZ dataset (English), then conduct fine-tuning with the parallel dialogs of different source languages.
  \item \textbf{DAML} In order to explore the relation between the parallel dialogs, we adopt DAML~\cite{Qian2019DomainAD} to train our model.
\end{itemize}

\subsection{Metrics}
Following \cite{Budzianowski2018MultiWOZA}, we adopt Inform Rate, Success Rate, and BLEU \cite{papineni-etal-2002-bleu} score as our main evaluation metrics. \textbf{Inform Rate} represents the accuracy of successfully providing the correct entity (e.g., the name of a restaurant that satisfies all user's constraints in the restaurant domain). \textbf{Success Rate} measures how well the system answers all the requested information. \textbf{BLEU} score is adopted to evaluate the quality of the generated response, compared with the ground truth response. We also use \textbf{Slot Accuracy} to evaluate the quality of dialog state tracking.

\begin{table*}[t]
\centering
\small
\begin{tabular}{llcccc}
\toprule
\textbf{Language} &\textbf{Model} & \textbf{BLEU}  & \textbf{Inform}  & \textbf{Success} & \textbf{Slot Accuracy} \\
\midrule
English &Vanilla  & $15.34_{\pm 1.90}$  & $43.00_{\pm 24.0}$ & $6.00_{\pm 2.83}$ & $71.07_{\pm 1.75}$  \\
&+English Pretrain  &  $20.39_{\pm 3.92}$  & $57.00_{\pm 12.73}$ & $11.00_{\pm 12.73}$ & $75.27_{\pm 8.19}$  \\
&DAML & $18.10_{\pm 0.26}$  &  $53.00_{\pm 4.24}$ & $5.00_{\pm 1.41}$ & $73.29_{\pm 0.10}$ \\

\midrule
French & Vanilla   & $16.93_{\pm 1.51}$ & $26.00_{\pm 2.83}$ & $5.00_{\pm 1.41}$ & $70.48_{\pm 2.86}$  \\
&+English Pretrain   & $19.94_{\pm 3.12}$    &  $23.00_{\pm 1.41}$ & $4.00_{\pm 0.00}$ & $73.74_{\pm 8.63}$   \\
&DAML & $18.07_{\pm 0.47}$  & $22.00_{\pm 2.83}$ & $5.00_{\pm 1.41}$ & $65.04_{\pm 6.81}$  \\

\midrule
Vietnamese &Vanilla  & $16.95_{\pm 1.09}$  & $23.00_{\pm 1.41}$ & $6.00_{\pm 2.83}$ & $68.46_{\pm 1.72}$  \\
&+English Pretrain   & $19.63_{\pm 1.15}$   & $25.00_{\pm 1.41}$ & $6.00_{\pm 2.83}$ & $76.36_{\pm 4.76}$   \\
&DAML & $18.89_{\pm 1.74}$ & $23.00_{\pm 1.41}$ & $5.00_{\pm 1.41}$ & $70.61_{\pm 8.36}$   \\

\midrule
Portuguese &Vanilla  & $13.35_{\pm 4.51}$  & $23.00_{\pm 1.41}$ & $5.00_{\pm 1.41}$ & $75.13_{\pm 5.16}$  \\
&+English Pretrain   & $14.13_{\pm 0.16}$   &  $24.00_{\pm 0.00}$ & $4.00_{\pm 0.00}$ & $69.68_{\pm 1.75}$   \\
&DAML & $16.66_{\pm 1.94}$  & $24.00_{\pm 0.00}$ & $6.00_{\pm 2.83}$ & $72.82_{\pm 2.68}$   \\

\midrule
Korean &Vanilla  & $7.29_{\pm 0.09}$ &   $24.00_{\pm 0.00}$ & $4.00_{\pm 0.00}$ & $69.93_{\pm 1.42}$  \\
&+English Pretrain   & $9.59_{\pm 1.17}$   & $25.00_{\pm 1.41}$ & $4.00_{\pm 0.00}$ & $78.04_{\pm 0.77}$  \\
&DAML & $8.15_{\pm 2.15}$ & $23.09_{\pm 1.41}$ & $4.00_{\pm 0.00}$ & $74.42_{\pm 1.92}$   \\

\midrule
Mandarin &Vanilla  & $3.42_{\pm 1.56}$  & $26.00_{\pm 2.83}$ & $5.00_{\pm 1.41}$ & $74.53_{\pm 4.98}$  \\
&+English Pretrain   & $7.13_{\pm 2.10}$  & $24.00_{\pm 0.00}$ & $6.00_{\pm 2.83}$ & $79.40_{\pm 0.25}$  \\
&DAML & $3.93_{\pm 0.29}$  & $24.00_{\pm 0.00}$ & $4.00_{\pm 0.00}$ & $69.83_{\pm 1.81}$  \\

\midrule
Hindi &Vanilla  & $15.24_{\pm 0.70}$  & $26.00_{\pm 2.83}$ & $5.00_{\pm 1.41}$ & $71.62_{\pm 2.36}$  \\
&+English Pretrain   & $16.37_{\pm 0.26}$    &  $23.00_{\pm 1.41}$ & $4.00_{\pm 0.00}$ & $75.79_{\pm 1.56}$  \\
&DAML & $14.55_{\pm 5.44}$  & $23.00_{\pm 1.41}$ & $6.00_{\pm 2.83}$ & $72.52_{\pm 1.44}$   \\

\midrule
Thai &Vanilla  & $12.36_{\pm 0.82}$  & $25.00_{\pm 1.41}$ & $6.00_{\pm 0.00}$ & $76.84_{\pm 2.36}$  \\
&+English Pretrain   & $10.76_{\pm 1.43}$   & $22.00_{\pm 2.83}$ & $4.00_{\pm 0.00}$ & $65.17_{\pm 0.66}$   \\
&DAML & $10.39_{\pm 6.71}$  & $25.00_{\pm 1.41}$ & $7.00_{\pm 4.24}$ & $77.15_{\pm 3.30}$   \\

\midrule
Average &Vanilla  & $12.61_{\pm 1.50}$  & $27.00_{\pm 3.89}$ & $5.25_{\pm 1.41}$ & $72.26_{\pm 1.81}$  \\
&+English Pretrain   & \textbf{14.75$_{\pm 1.63}$}   &  \textbf{27.88$_{\pm 1.24}$} & \textbf{5.38$_{\pm 2.30}$} & \textbf{74.18$_{\pm 2.77}$}   \\
&DAML &  $13.59_{\pm 2.26}$ & $27.13_{\pm 1.24}$ & $5.25_{\pm 1.41}$ & $71.96_{\pm 0.04}$   \\

\bottomrule
\end{tabular}
\caption{\label{citation-guide}
Performances of three benchmark models in terms of BLEU score, Inform Rate, Success Rate and Slot Accuracy for each language.
}
\end{table*}

\subsection{Results}

The results of all three benchmark models for each language are included in the Table \ref{citation-guide}. From the table, we observe that pre-training on English MultiWOZ corpus improves the BLEU score for 7 out of 8 languages, and increases the slot accuracy for 6 out of 8 languages. With the English pre-training, the model does not only perform better on the dialog state tracking task, but also better on the language generation task. The improvement for non-English language indicates that dialog knowledge from the English pre-training data can be adapted to a new language through paralleled dialog data. Therefore, the similar structures that different languages share help the model to generalize to new languages based on the embedded information about English data. Thai is the only language that ``Vanilla+English Pretrain'' setting does not perform as well as the ``Vanilla'' model, not only for BLEU score, but also for slot accuracy. This results from the differences of dialog utterances and structures between Thai and English.

The DAML approach, without introducing extra English corpus, improves BLEU score for 6 out of 8 languages, and improves slot accuracy for 5 out of 8 languages. By forcing the model to learn the similar structures that different languages share, the DAML approach outperforms the model pre-trained on English corpus for both Thai and Portuguese. The performance of DAML over the ``Vanilla'' setting also shows that parallel corpus brings significant advantages when the pre-trained multilingual models are used for downstream tasks in a language that we do not have a lot of available data.


\section{Conclusion}
We created a new multilingual dialog data with eight languages focusing on customer service tasks. 
We find that our model, which uses meta-learning to learn the shared structures between languages, performs significantly better than normal training in a few-shot setting and could achieve comparable result when there is enough training data. 

In future work, we plan to expand the dataset to 30 languages. In addition, we will study how to perform zero-shot generation on all languages, and how to improve performance on both tasks and generations in the zero-shot setting. 



\bibliography{anthology, custom}
\bibliographystyle{acl_natbib}

\end{document}